# Automatic Detection of Node-Replication Attack in Vehicular Ad-hoc Networks

Mohammed GH. I. AL Zamil

Department of Computer Information Systems

Faculty of Information Technology and Computer Sciences

Yarmouk University

Mohammedz@yu.edu.jo

Tel: +962 777 260 802

## Abstract

Recent advances in smart cities applications enforce security threads such as node replication attacks. Such attack is take place when the attacker plants a replicated network node within the network. Vehicular Ad hoc networks are connecting sensors that have limited resources and required the response time to be as low as possible. In this type networks, traditional detection algorithms of node replication attacks are not efficient. In this paper, we propose an initial idea to apply a newly adapted statistical methodology that can detect node replication attacks with high performance as compared to state-of-the-art techniques. We provide a sufficient description of this methodology and a road-map for testing and experiment its performance.

# 1. Introduction

Modern technologies are deeply dependent on cordless communication, that is mostly based on wireless sensors [12, 13, 14, 25]. Sensors are edge computing devices, which has limited resources (memory, power, computation) and dedicated to achieving a specific task. Recent advances in vehicular ad hoc networks suggest the application of V2X in which vehicles can communicate on roads in two forms: Vehicles-to-Vehicles and Vehicles-to-Infrastructure. Such communication paradigm is prone to different security attacks, since wireless signals are sent and received in an open medium [19, 20].

In the literature, wireless communication attacks are classified into two different kinds: internal and external attacks [1, 2]. Internal attacks are compromised by the internal network nodes that used to discover secured encrypted information, report false alarms (such as accidents), breach routing attacks, and many others. On the other hand, the external attacks are compromising attacks from sources that are not belonging to the network itself; such as deploying fake nodes with power resources [15, 17, 18, 21].

One of the most destructive attacks that V2X technology suffer is the node replication attack. It is the situation in which the adversary node replicates original ones with features that make it almost the same. The supplicated node, then, acts as part of the network itself. The network controller, then, cannot distinguish it from other trusted nodes. Therefore, it can manipulate the neighbouring nodes by sending fake information [22, 23].

The replicated node has the power to imitate the identifier of the original node with the network; appearing as an original node. Such attack is used to distribute false information, changing the communicated data, making warm-whole attack, and removing selected packets [16, 24].

Existing techniques for preventing such attacks are too expensive in terms of the need for computation power, memory, and perfect synchronization. All these resources are hard to be available in V2X environments. Therefore, there is a need to develop new techniques that benefit from the Vehicle-to-Infrastructure paradigm to prevent such attacks.

This paper introduces previous work on attacking this problem; highlighting a methodology that initiate an idea that could be useful in this domain. The probabilistic spatial technique that is proposed in this paper is useful and easy to implement, but requires extensive empirical experiments to be proved.

This paper is organized as follows: the first section introduces the problem and its applicable domain. Section 2 illustrates the related work in the literature with a comparison with our proposed methodology. Section three describes the methodology and the proposed protocol.

## 2. Literature Review

Node-replication attacks in wireless sensor networks have been classified, in the literature, according to algorithms' localization and the nature of the objects within a WSN. However, Centralized algorithms designed to communicate with a central unit that keeps tracking objects within the network and, thus, responsible for detecting replication attacks. On the other hand, many algorithms have been designed to distribute the detection task over some nodes within the network (sometimes called witnesses).

Centralized techniques such as random key pre-distribution algorithm [3], SET [4], and Fast-Detection [5] provide high performance methodologies to detect node-replication attack. The random key pre-distribution relies on the central station that assigns a key to every node joined the network. The central station conducts a protocol to track nodes and their behavior. Unfortunately, such technique guarantees tracking keys; not nodes. Choi et al. proposed SET; an efficient method that depends on applying set-operations that allows central station to detect replication using simple set-theory operation such as intersection.

To handle mobility, in which nodes often change their location, Ho et al. proposed a fast detection algorithm that relies on sequential probability ratio test. During the configuration, a maximum speed is defined. Once the speed of two consecutive claims exceeds the maximum configured speed, the node is considered replicated.

Recently, modern WSNs (such as vehicular ad hoc networks VANETs) are ad hoc where no predefined information (state) is kept about nodes. Distributed algorithms to detect node-replication attacks received noticeable concerns from specialists. For this reason, many distributed algorithms have been proposed to handle these attacks.

Real-time detection [6] has been proposed to provide a semi-distributed algorithm. The idea behind this technique is to verify the fingerprint of nodes by the central station (Centralized) and the neighboring nodes (distributed).

Broadcasting nodes within cordless networks has been discussed in [7]. This approach based on broadcasting the IDs of nodes within the network to every neighboring node. Furthermore, the broadcaster must declare its location as well. Once a node receives such information, it must apply a comparison with the other nodes' list of neighbors. If there exists collusion among IDs in at least two nodes with different locations, the corresponding one is revoked. Such solution seems to be perfect, but in our environment, it requires high communication and may overload the network as the limited computation power is a critical issue.

Parono et al. [8] have suggested a distribution detection protocol that is able to detect node replication attack in a wireless environment. The proposed Randomized and line-Selected Multicast protocol is relying on the assumption that the geographic coordination (location) of every node in the wireless network is well-known by every node in the network. Moreover, it assumes that the signature (public key) of each node is known as well. Such solution is functioning well; especially when base-units in these networks are treated as trusted entities. The core disadvantage of this solution is the ability of the adversary to attack the trusted unit itself (base-unit).

Defending the node replication attacks has been suggested by Dutertre et al. [9] using par-wise key establishment protocol. The researchers argued that such methodology provides a defending wall to replication attacks; to some degree. It assumes that the deployment of nodes is done based on time; there is a maximum number of nodes that can be deployed in a certain time. Such approach classifies deployment of nodes as a set of generations. For each generation, a master key is defined to distinguish nodes of the same generation. Once a node is joined a specific generation, it should remove this key so that replicated node cannot recognize it.

Finally, another interested approach is proposed by Liu et al. [11]. It is based on the distribution of location-aware key. Such protocol is distinguishing among the types of node. It defines a service node that is elected to be a server of trusted keys. Incoming nodes must communicate with the service node to declare their location and keys. Once a node declares its location that is a replacement of another existing node, the service node detects the replica and prevents it from joining the network.

While the last two solutions require high computational power, they proved their performance in terms of detecting replication attacks. Thus, spatial-based solutions are much powerful than traditional ones. Therefore, there is a need to upgrade these solutions with algorithms and techniques that are required low computational power and resources.

## 3. Methodology

This section illustrates our proposed methodology. At initial stage, the proposed technique initiates a list of meta-information about nodes that are connected to each central base station. Then, a description of the detection protocol is described that could be adapted at execution time.

### 3.1 Initialization

Once a node enters the range of a given central-station (for example, a road), it generates and assign a key to the vehicle. The key is computed according to the symmetric variant method described in [10] as follows:

$$P(x,y) = \sum_{i,j=0}^{t} (a_{ij} \times x^i y^j) \, mod \, Q$$

Where $GPS_i$ is the position of a node, $t_i$ is the polynomial degree, $G_i$ is the group membership ID, and $Pr_i$ is the permission-level. Furthermore, $1 \leq a_{ij} \leq Q - 1$. For new incoming vehicle to join, the central-station retrieves all nodes $u \in G_i$ and computes the identity as follows:

$$f_i(y) = P(x = i, y) = f(H(i|Key_i), y) = \sum_{i,j=0}^{t} (a_{ij} \times [H(i|Key_i]^i \ y^i \ mod \ Q$$

To clarify these formulas, suppose we have a group of (t) nodes that requested keys. The node with the largest identity in a given group is chosen. The group leader, then, send this value to the other (t-1). The central-station computes the keys using P(x, y) function. For new incoming node, it has to pick a key from some neighboring node and compute the key using (fi(y)).

## 3.2 Protocol Description

For new incoming node to know their previous ancestors (old nodes' keys), it must seek for the highest deployed generations. Initially, the central station keeps states for previous generated keys. For the first generation, the central station sets the parameter (t=0). Such parameter would be useful for timely synchronization in case of some applications. To differentiate among different groups, a period is clearly defined among consecutive generations. This mechanism allows central-station to retrieve exact nodes for each group during initialization phase. For a group of nodes that is belong to the same temporal set of nodes that request a key, the service node should make sure that the new node is an original one by comparing the current time, time of deployment, and the period time.

Suppose that a given node (n) tries to connect with some neighboring node by sending "Hello" message. The receiver would response according to the generation level of sending node. If the sending node is newly deployed or its key-time is still valid (Not Expired), the verification will be failed implying that the sending node is already deployed. Otherwise, if the sending node succeeded the verification, the receiver will compute the key using equation (2) and send it back to the sender node.

## 4. Experiment and Results

In this section, two experiments are discussed to show the efficiency of the proposed protocol, the first highlights a comparison in terms of communication and memory complexity. The second experiment highlights the detection rate as compared to state-of-the-art techniques.

## 4.1 Communication Cost

Suppose that existing network consists of (n) nodes, in which the degree of neighbors is indicated as (D) and the diameter of the network communication is depicted as (s). Table 1 shows a comparison among state-of-the-art techniques.

Table 1 Comparison among different techniques

| Protocol | Communication Complexity | Memory Complexity |
|---|---|---|
| EDD | $O(n)$ | $O(n)$ |
| SDC | $O(n) + O(s)$ | NAP |
| Randomized Multicast | $O(n^2)$ | $O(n)$ |
| SET | $O(n)$ | $O(D)$ |
| RED | $O(\sqrt{n})$ | $O(D)$ |

The proposed protocol in this paper achieves $O(n)$ communication complexity and $O(s)$ memory complexity. Therefore, the performance of the proposed protocol is acceptable as compared to existing ones.

## 4.2 Detection Rate

As shown in Figure 1, the detection rate of the proposed technique (PPP) achieved higher performance at different network load.

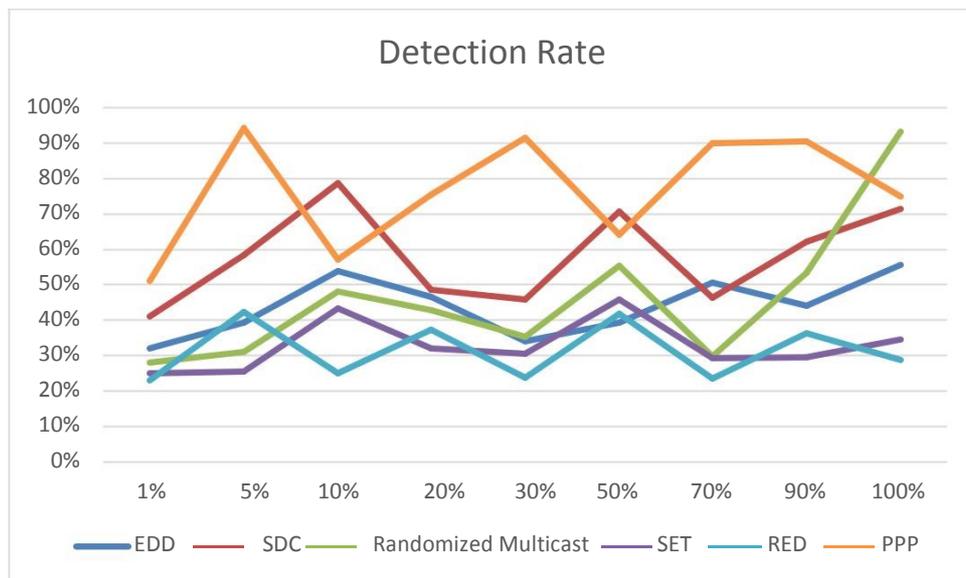

Figure 1 Detection Rate

At high network load, PPP outperformed all techniques except the randomized multicast protocol. While at average network load, there was a competition between SDC and PPP.

# 5. Conclusion

This paper introduces a node replication detection technique based on a statistical model. The paper showed the methodology to implement the proposed protocol with high performance in terms of computation, communication, and memory complexity. Furthermore, the proposed protocol achieved above average detection rate as compared to state-of-the-art techniques.